\documentclass[10pt, conference]{IEEEtran}

\usepackage[latin1]{inputenc}  
\usepackage{graphicx,url}
\usepackage[T1]{fontenc}
\usepackage{multirow}  
\usepackage{hhline}
\usepackage{graphicx}
\usepackage{subfigure,amssymb,amsmath}
\usepackage{color,soul}
\usepackage[dvipsnames]{xcolor}
\usepackage{bbm}
\usepackage{float}
\usepackage{color}
\makeatletter  
\newif\if@restonecol  
\makeatother

\usepackage[linesnumbered,ruled,vlined]{algorithm2e}
\usepackage{algpseudocode}  
\usepackage{amsmath}  
\usepackage{fancyhdr} 
\usepackage{tabularx}
\usepackage{comment} 

\renewcommand\thetable{\arabic{table}}
\IEEEoverridecommandlockouts
\IEEEpubid{\makebox[\columnwidth]{978-3-903176-31-7~\copyright~2020 IFIP\hfill} \hspace{\columnsep}\makebox[\columnwidth]{ }}

\usepackage{etoolbox}
\makeatletter
\patchcmd{\@makecaption}
  {\scshape}
  {}
  {}
  {}
\makeatother

\begin{document}


\title{Online Feature Selection for Efficient Learning \\
in Networked Systems}

\author{\IEEEauthorblockN{Xiaoxuan Wang  and  Rolf Stadler}

 \IEEEauthorblockA{
Dept. of Computer Science, Digital Futures
\\
KTH Royal Institute of Technology, Sweden
 }
  \newline
Email: \{xiaoxuan, stadler\}@kth.se
\\
\today
}

\maketitle

\thispagestyle{plain}
\pagestyle{plain}

\begin{abstract}
\label{sec:abstract}
Current AI/ML methods for data-driven engineering use models that are mostly trained offline. Such models can be expensive to build in terms of communication and computing cost, and they rely on data that is collected over extended periods of time. Further, they become out-of-date when changes in the system occur. To address these challenges, we investigate online learning techniques that automatically reduce the number of available data sources for model training. We present an online algorithm called Online Stable Feature Set Algorithm (OSFS), which selects a small feature set from a large number of available data sources after receiving a small number of measurements. The algorithm is initialized with a feature ranking algorithm, a feature set stability metric, and a search policy. We perform an extensive experimental evaluation of this algorithm using traces from an in-house testbed and from a data center in operation. We find that OSFS achieves a massive reduction in the size of the feature set by 1-3 orders of magnitude on all investigated datasets. Most importantly, we find that the accuracy of a predictor trained on a OSFS-produced feature set is somewhat better than when the predictor is trained on a feature set obtained through offline feature selection. OSFS is thus shown to be effective as an online feature selection algorithm and robust regarding the sample interval used for feature selection. We also find that, when concept drift in the data underlying the model occurs, its effect can be mitigated by recomputing the feature set and retraining the prediction model. 

\end{abstract}


\section{Introduction}
\label{sec:introduction}

Data-driven network and systems engineering is based upon applying AI/ML methods to data collected from an infrastructure in order to build novel functionality and management capabilities. This is achieved through learning tasks that are trained on this data. Examples are KPI prediction and forecasting through regression and anomaly detection through clustering techniques.

The way AI/ML methods are currently applied in data-driven engineering has several drawbacks. For instance, model training is often expensive in terms of communication and computing costs, and it relies on data that must be collected over extended periods of time. Second, since training is performed offline, models generally become out-of-date and the performance of learning tasks degrades after changes in the system or in the environment. To address these challenges, we advocate investigating online learning techniques and performing model re-computation after detecting a change in the data distribution underlying the model.

Creating a model for a learning task, for example a model for KPI prediction, includes (1) the extraction of metrics from devices in a network or an IT system, (2) the transfer of this data to a processing point, (3) the pre-processing of this data (removing outliers,  ), and (4) the training  of the model. The first two steps are part of the monitoring process. Each of these four steps incurs computing and/or communication costs that increase at least linearly with the number of sources from which data are extracted and with the number of observations that are used for training the model.

This paper focuses on automatically reducing the number of data sources involved in creating an effective model with the goal of significantly reducing monitoring cost and model training cost. We propose a novel online source-selection method that requires only a small number of measurements to significantly reduce the number of data sources needed for training models that are effective for learning tasks. 

Using the terminology of machine learning, we call a (one-dimensional, scalar) data source also \emph{a feature}, and we refer to measurements taken from a set of data sources at a specific time as \emph{a sample}.

Our approach consists of (1) ranking the available data sources using (unsupervised) feature selection algorithms and (2) identifying \emph{stable feature} sets that include only the top $k$ features. We call a feature set stable if it remains sufficiently similar when additional samples are considered. 

We present an online algorithm, which we call Online Stable Feature Set Algorithm (OSFS). It selects a small feature set from a large number of available data sources using a small number of measurements, which allows for efficient and effective learning. The algorithm is initialized with a feature ranking algorithm, a feature set stability metric, and a search policy. 

We perform an extensive experimental evaluation of this algorithm using traces from an in-house testbed and the FedCSIS Challenge dataset \cite{janusz2020network}. We find that OSFS achieves a massive reduction in the size of the feature set by 1-3 orders of magnitude on all investigated datasets. The number of samples needed to compute the feature sets averages around 430, which means that OSFS produces a stable feature set within an average of 7 minutes on the KTH testbed, when monitoring data is collected very second. Most importantly, we find that the performance of a predictor trained on a OSFS-produced feature set is similar to the one that uses offline feature selection. OSFS is thus shown to be effective as an online feature selection algorithm and robust regarding the sample interval used for feature selection.

We also find that, if concept drift occurs, its effect can be effectively mitigated by recomputing the feature set and retraining the prediction model.


Our results suggest that many data-driven functions in networked systems can be trained rapidly and with low overhead if they are re-trained after a change is detected. In particular, they do not require a lengthy monitoring phase and expensive offline training before prediction can begin. 

This paper is a significant extension of earlier work published in CNSM2020 \cite{wang2020online}. In this paper, we present OSFS in a new, generic form that allows instantiation with feature ranking algorithm, feature stability metric and search policy. We introduce and evaluate an additional stability metric which can be used with OSFS. The evaluation in this paper is more in-depth and more extensive. It includes studying the performance of OSFS on an additional dataset, namely, the FedCSIS challenge dataset. Also, we study the effect of concept drift on prediction accuracy and how to mitigate the increasing error through feature set re-computation and model re-training. Finally, we revise and extend the related work section.

The rest of the paper is organized as follows. Section \ref{sec:problem_formulation} formulates the problem we address in the paper. Section  \ref{sec:creating_ranked_feature_lists}
describes the feature selection methods we use to obtain ranked feature lists. Section \ref{sec:Computing_stable_feature_sets} introduces the concept of the stable feature set. Section \ref{sec:online_feature_selection_with_low_overhead} presents our online features selection algorithm OSFS. Section \ref{sec:change_detection_and_model_recomputation} discusses change detection and model rec-computation. Section \ref{sec:testbed} details our testbed, the traces we generate from this data and the FedCSIS challenge data set. Section \ref{sec:evaluation_of_OSFS_on_different_data_sets} contains the evaluation results. Section \ref{sec:related_work} surveys related work. Finally, Section \ref{sec:conclusions} presents the conclusions and future work.

\section{Problem formulation and approach}
\label{sec:problem_formulation}

We consider a monitoring infrastructure that collects readings from a set $\emph{\textbf{F}}$ of $n$ distributed data sources (or features). Each feature has a one-dimensional, numerical value that changes over time. We collect readings at discrete times $t$ and store them in sample vectors $\emph{\textbf{X}}_t \in \mathbb{R}^n,  t=1,2,3,..$ . Our objective is to identify a subset $\emph{\textbf{F}}_{k} \subset \emph{\textbf{F}}$ with $k \ll n$ features using the samples $\emph{\textbf{X}}_1, \emph{\textbf{X}}_2, ..,  \emph{\textbf{X}}_{t_k}$. Second, we consider a learning task, like KPI prediction or anomaly detection, whose model is trained using the samples $\emph{\textbf{X}}_t \in \mathbb{R}^k$ with the features from  $\emph{\textbf{F}}_{k}$.
 
\begin{figure}[!ht]
  \centering
  \includegraphics[width=0.97\linewidth]{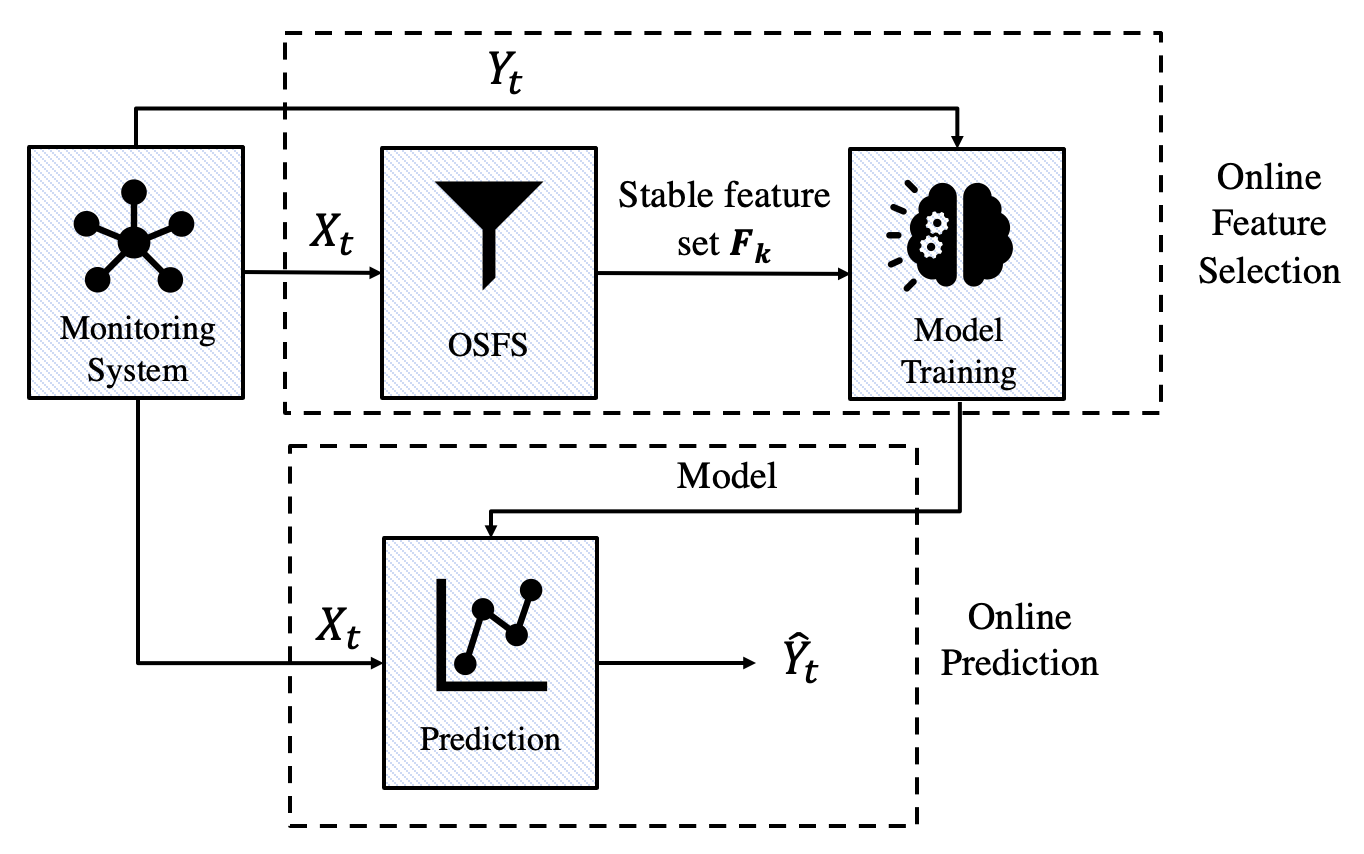}
  \caption{Online learning: feature selection and prediction. OSFS reads a stream of samples ($\emph{\textbf{X}}_t$) and extracts a stable feature set $\emph{\textbf{F}}_{k}$.}
  \label{fig:online_learning}
\end{figure}

Figure \ref{fig:online_learning} illustrates the process how the subset $\emph{\textbf{F}}_{k}$ is created. Central to the process is the algorithm OSFS which will be introduced in Section \ref{sec:online_feature_selection_with_low_overhead}. 
 
In order to keep the monitoring overhead low for collecting the samples and the computational overhead for training the model associated with the learning task, the numbers for $k$ and $t_k$ should be small. Note that $k$ indicates the number of data sources that need to be monitored to train the model and $t_k$ refers to the number of measurements that are needed to compute $\emph{\textbf{F}}_{k}$. Assuming periodic readings, $t_k$ further indicates the time it takes until the feature set $\emph{\textbf{F}}_{k}$ is available.

Our goal thus is to select $k$ and $t_k$ as small as possible, while enabling the models trained using $\emph{\textbf{F}}_{k}$ to be effective.

The task of selecting a subset of features from a larger set is called \emph{feature selection}  in machine learning and data mining and is a well-studied topic area.
(See Section \ref{sec:related_work}). 
We are specifically interested in unsupervised feature selection methods, whereby the values of the target are not known during the feature selection process, i.e., the process to compute $\emph{\textbf{F}}_{k}$. This allows us to keep the feature selection process independent from the learning task and will enable different learning tasks in a system to share the same feature subset.

The specific problem we address in this paper is to design an online algorithm that reads a sequence of $n$-dimensional sample vectors $\emph{\textbf{X}}_1, \emph{\textbf{X}}_2, .. $ one by one, computes $k$ and the feature set $\emph{\textbf{F}}_{k}$, and terminates after step $t_k$. The values for $k$ and $t_k$ must be small, while $\emph{\textbf{F}}_{k}$ must be equally effective as the feature set selected by the same ranking algorithm in an offline setting on the complete data stream. We measure the effectiveness of a feature set by computing the prediction error of the predictor trained with the feature set.

In our approach, we choose an unsupervised feature selection method that ranks the n features at every step $t$ of the online algorithm and checks how the top $i$ features ($i=1,..,n$) change with increasing $t$. We introduce two metrics to measure the stability of the feature set in Section \ref{sec:Computing_stable_feature_sets}. If the feature set of the top $k$ features has become sufficiently stable, the algorithm terminates and the values for $\emph{\textbf{F}}_{k}$, $k$ and $t_k$ are returned (see Section \ref{sec:online_feature_selection_with_low_overhead}).  

Note that we assume here that the feature set $\emph{\textbf{F}}$ is fixed. In a real system that runs over some time, changes to the physical configuration or the virtualization layer occur, which result in changes to the set of available measurement points, i.e. the feature set. In such a case, the online algorithm must be re-started. Note also that we do not investigate how many samples are needed to train an accurate model for the learning task. This is left for future work.

\section{Creating ranked feature lists}
\label{sec:creating_ranked_feature_lists}

In this section, we describe two algorithms (ARR, LS) that produce a ranked feature list from a list of samples. They are based on unsupervised feature selection methods from the literature. 
We will evaluate the suitability of these algorithms for our online feature selection method OSFS in Section \ref{sec:online_feature_selection_with_low_overhead}. 

Table \ref{notation} shows the notation we use in the paper. The available data for computing the feature set $\emph{\textbf{F}}_{k}$ is presented as a design matrix $\emph{\textbf{X}} \in \mathbb{R}^{m\times n}$, whose $n$ columns represent the feature vectors and $m$ rows represent the samples in the data set. Since we assume that the samples arrive in sequence one-by-one, $m$ is increasing over time and can be interpreted as a time index. 

\begin{table}[ht]
    \centering
    \caption{Table of notation}
    \label{notation}
    \scalebox{0.9}{
    \begin{tabular}{|c|c|}
    \hline
    $\emph{\textbf{X}}$& data set \\
    \hline 
    $n$& number of features in $\emph{\textbf{X}}$ \\
    \hline 
    $m$& number of samples in $\emph{\textbf{X}}$\\
    \hline 
     $\emph{\textbf{X}}_{i,:}\ $or $\emph{\textbf{X}}_{i}    (i=1,...,m)$ & $i$-th row or $i$-th sample of $\emph{\textbf{X}}$\\
    \hline
    $\emph{\textbf{X}}_{:,j}\ (j=1,...,n)$& $j$-th column or $j$-th feature vector of $\emph{\textbf{X}}$\\
    \hline
    $\mathit{X_{i,j}}$& element of the $i$-th row and the $j$-th column of $\emph{\textbf{X}}$\\
    \hline
    $k$& number of selected features \\
    \hline 
    $t_k$& number of samples used for feature selection\\
    \hline
    $\emph{\textbf{F}}$& set of all available features\\
    \hline
    $f_{j}\ (j=1,...,n)$& single feature in $\emph{\textbf{F}}$\\
    \hline 
    $\emph{\textbf{F}}_{k}$& subset with $k$ selected features\\ 
    \hline
    $\emph{\textbf{F}}_{k,t}$& the top $k$ features computed on $t$ samples\\
    \hline
    \end{tabular}}
\end{table}

The first algorithm is Adapted Relevance Redundancy Feature Selection (ARR). It is based on the Relevance Redundancy Feature Selection (RRFS) method \cite{ferreira2012}, which we adapted to compute a ranked feature list. ARR uses two criteria in assessing the rank of a feature: (a) relevance, which relates to the distance of a feature vector to the mean of all feature vectors, and (b) redundancy, which relates to the cosine similarity between a feature vector and the vectors of all other features. High relevance and low similarity result in a high score. 
The pseudo-code of ARR is given in Algorithm \ref{arr_algorithm}.  First, the relevance of a feature is computed as the mean absolute difference of its feature vector from the mean (line 3). Then, the relevance of a feature is computed as the sum of the cosine similarity of its feature vector and each feature vector of the data set (lines 5-6). The score for ranking a feature is the relevance value divided by the redundancy value (line 7). The computational complexity of ARR is $O(n^2m)$.
Recall that the cosine similarity between two features vectors $\emph{\textbf{X}}_{:,a}$ and $\emph{\textbf{X}}_{:,b}$ is calculated as:
\begin{equation}
    cosim(\emph{\textbf{X}}_{:,a},\emph{\textbf{X}}_{:,b})=\left | \frac{\sum_{i=1}^{m}(\mathit{X_{i,a}}\mathit{X_{i,b}})}{(\sqrt{\sum_{i=1}^{m}\mathit{X_{i,a}}^{2}})(\sqrt{\sum_{i=1}^{m}\mathit{X_{i,b}}^{2}})}\right |
\end{equation}

\begin{algorithm} 
  \label{arr_algorithm}
  \caption{Adapted Relevance Redundancy Feature Selection (ARR)}  
  \KwIn{Data matrix $\emph{\textbf{X}} \in \mathbb{R}^{m\times n}$}  
  \KwOut{Ranked feature list $\emph{\textbf{F}}^{'}$}
  $\emph{\textbf{F}}^{'}$=[]\;
  \For{$i=1;i\leq n;i++$} 
  {  
    $relevance_{i}=\sum_{j=1}^{m}\left | \mathit{X_{j,i}}-\overline{\emph{\textbf{X}}_{:,i} }\right |$\;  
    $sim\_sum_{i}=0$\;  
    \For{$j=1;j\leq n;j++$}  
    {  
      $sim\_sum_{i}+=cosim(\emph{\textbf{X}}_{:,i},\emph{\textbf{X}}_{:,j})$\;  
    }  
    $score_{i}=\frac{relevance_{i}}{sim\_sum_{i}}$\;
  } 
   Construct $\emph{\textbf{F}}^{'}$ as a list of all features sorted by $score_{i}$ in descending order\;
  return $\emph{\textbf{F}}^{'}$\;  
\end{algorithm}  

The second algorithm is Laplacian Score (LS) \cite{he2006laplacian}. It follows the so-called filter method which examines intrinsic properties of the data to evaluate the features. LS ranks those features high that preserve locality with respect to a neighborhood graph. Algorithm 2 shows the pseudocode of LS. Input parameters for this algorithm are the design matrix and the number of local neighbors $K$.
First, using the $m$ sample vectors as the nodes of the graph, a neighborhood graph is constructed with the $K$ nearest neighbors of each node as the links of the graph (line 1). Then, the graph connectivity and the distance between node pairs are used to compute the weight matrix (line 2). The graph Laplacian matrix is computed in line 4. The Laplacian score of all $n$ features is obtained in the for loop (lines 5-7). A low score for a feature signifies high locality preservation (see \cite{he2006laplacian} for justification), and the features are ranked according to increasing score (line 8). The computational complexity of the LS algorithm is $O(nm^2)$.

In the evaluations reported in Sections V and VI of this paper, we choose $K$ in relation to the value of $m$: $K=2$ for $0<m\leq 16$; $K=5$ for $16<m\leq 128$; $K=10$ for $m>128.$

\begin{algorithm} 
  \label{ls_algorithm}
  \caption{Laplacian Score (LS)}  
  \KwIn{Data matrix $\emph{\textbf{X}} \in \mathbb{R}^{m\times n}$, $K \in \mathbb{N}$ nearest neighbors}  
  \KwOut{Ranked feature list $\emph{\textbf{F}}^{'}$}
  
  Construct graph $\emph{\textbf{G}}$ of $K$ nearest neighbors from nodes $\emph{\textbf{X}}_{i,:}, i=1,..,m$\;
  Compute weight matrix $\emph{\textbf{S}} \in \mathbb{R}^{m\times m}$ from $\emph{\textbf{G}}$
  \begin{equation*}
  S_{ij}=\left\{\begin{matrix}
   e^{-\left \| \emph{\textbf{X}}_{i,:}-\emph{\textbf{X}}_{j,:} \right \|^{2}} & if\ nodes \ i \ and \ j \ are\ connected,\\ 
   0& otherwise
    \end{matrix}\right.
    \end{equation*}\\
  $\emph{\textbf{D}}= diag(\emph{\textbf{S1}})$ where 
  $\emph{\textbf{1}}=\left [ 1,...,1 \right ]^{T}$,
  $\emph{\textbf{D}}\in \mathbb{R}^{m\times m}$\;
  $\emph{\textbf{L}}= \emph{\textbf{D}} - \emph{\textbf{S}}$\;
  
  \For{$i=1;i\leq n;i++$} 
  {  
    $\emph{\textbf{V}}_{i}=\emph{\textbf{X}}_{:,i}-\frac{\emph{\textbf{X}}_{:,i}^{T}\emph{\textbf{D}}\emph{\textbf{1}}}{\emph{\textbf{1}}^{T}\emph{\textbf{D}}\emph{\textbf{1}}}\emph{\textbf{1}}$ where
    $\emph{\textbf{V}}_{i}\in \mathbb{R}^{m\times 1}$\;  
    $lscore_{i}=\frac{\emph{\textbf{V}}_{i}^{T}\emph{\textbf{L}}\emph{\textbf{V}}_{i}}{\emph{\textbf{V}}_{i}^{T}\emph{\textbf{D}}\emph{\textbf{V}}_{i}}$\;
  } 
   Construct $\emph{\textbf{F}}^{'}$ as a list of all features sorted by $lscore_{i}$ in ascending order\;
  return $\emph{\textbf{F}}^{'}$\;  
\end{algorithm}

\section{Computing stable feature sets}
\label{sec:Computing_stable_feature_sets}

In Section \ref{sec:creating_ranked_feature_lists} we discussed the algorithms ARR and LS, which produce ranked feature lists after reading $m$ samples $\emph{\textbf{X}}_{1}$, ..., $\emph{\textbf{X}}_{m}$. In an online setting, where samples become available one-by-one at discrete times $t=1,2,...$, the value of $m$ can be interpreted as time. Let's assume we run a feature ranking algorithm at time $t_1$ and compute the set $\emph{\textbf{F}}_{k, t_1}$ with the top $k$ features. This set will generally be different from the set $\emph{\textbf{F}}_{k, t_2}$ produced by the same algorithm at a later time $t_2$ for statistical reasons. Using the standard assumption from statistical learning that samples are drawn from static distributions \cite{vapnik98}, we expect that the sequence $\emph{\textbf{F}}_{k, t}$ converges to a set $\emph{\textbf{F*}}_{k}$ with increasing time $t$. Our objective for this section is to identify heuristic criteria that determine at which time the sequence $\emph{\textbf{F}}_{k, t}, t=1,2,3,...$ has sufficiently converged or, as we also say, $\emph{\textbf{F}}_{k, t}$ has become ``stable''.

We define two criteria for feature set stability, which use frequency-based stability methods \cite{nogueira2017stability}. The first is based on the similarity of two feature sets, the second on the variance of feature frequencies in feature sets. 

\subsection{A stability condition based on set similarity}
\label{subsec:feature_set_similarity}
Given two feature sets $\emph{\textbf{F}}_{k, t_1}$ and $\emph{\textbf{F}}_{k, t_2}$ ($k\in \mathbb{N}$), we define the similarity between them as the fraction of joint features:
\begin{equation}
    sim(\emph{\textbf{F}}_{k, t_1},\emph{\textbf{F}}_{k, t_2}):=|\emph{\textbf{F}}_{k, t_1}\cap \emph{\textbf{F}}_{k, t_2}|/k
\end{equation}
The values of the metric $sim$ lie between 0 and 1. $0$ means no similarity, i.e. $\emph{\textbf{F}}_{k, t_1}$ and $\emph{\textbf{F}}_{k, t_2}$ have no common features, 
while $1$ means the maximum similarity, i.e. $\emph{\textbf{F}}_{k, t_1}$ and $\emph{\textbf{F}}_{k, t_2}$ are identical.

\begin{figure}[ht]
\centering

\subfigure[$sim(\emph{\textbf{F}}_{k,t/2},\emph{\textbf{F}}_{k,t})$ vs $t$ in ARR]{
\begin{minipage}[t]{0.48\linewidth}
\centering
\includegraphics[width=1.7in]{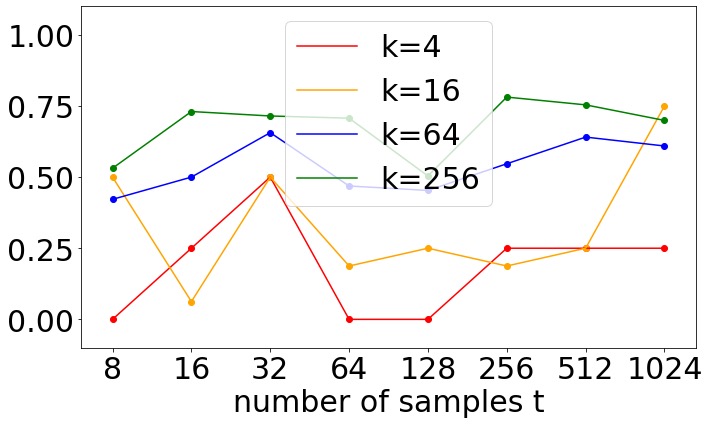}
\end{minipage}%
}%
\subfigure[$sim(\emph{\textbf{F}}_{k,t/2},\emph{\textbf{F}}_{k,t})$ vs $t$ in LS]{
\begin{minipage}[t]{0.48\linewidth}
\centering
\includegraphics[width=1.7in]{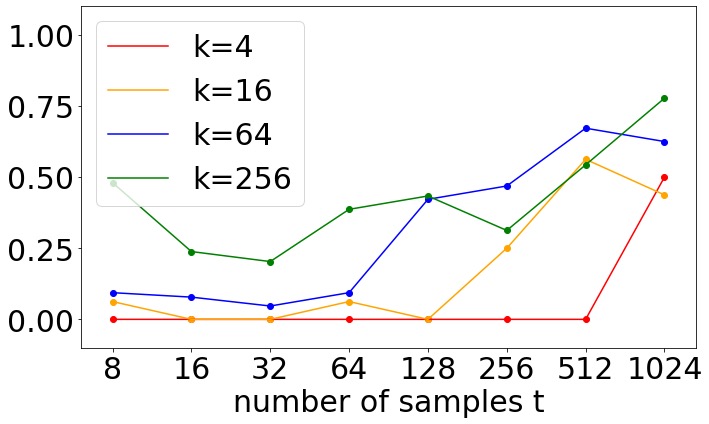}
\end{minipage}%
}%

\centering
\caption{The similarity of consecutive feature sets over time using ARR and LS with samples from the KV flash-crowd data set and start point $t=1$.}
\label{pic:stability_s}
\end{figure}

When we study the evolution of $sim(\emph{\textbf{F}}_{k,t},\emph{\textbf{F}}_{k, 2t})$ using the traces described in Section \ref{sec:testbed}, we observe that the similarity metric tends to grow with the increase of $k$ and $t$. Also, this effect becomes stronger the larger $k$ and $t$ are. Figure \ref{pic:stability_s} shows an example with data from the KTH testbed running a KV service.  

We introduce the first heuristic notion of a \emph{stable feature set}: 
\begin{equation}
    \label{equ:sim_con}
    \begin{split}
    \emph{\textbf{F}}_{k,t}\ is\ stable,\ iff\   sim(\emph{\textbf{F}}_{k,t},\emph{\textbf{F}}_{k,2t}) > \eta \\
    \ and\ sim(\emph{\textbf{F}}_{k,t},\emph{\textbf{F}}_{k,2t}) > sim(\emph{\textbf{F}}_{k,2t},\emph{\textbf{F}}_{k,4t})
    \end{split}
\end{equation}

Based on our experience with operational data, we set $\eta=0.5$ for this work. This means that $\emph{\textbf{F}}_{k,t}$ must share at least half of its features with $\emph{\textbf{F}}_{k,2t}$, which is computed with double the number of samples. The second condition in equation (\ref{equ:sim_con}) is met if the sequence of similarity values reaches a local maximum. In Figure \ref{pic:stability_s} (a) $\emph{\textbf{F}}_{64,32}$ and in Figure \ref{pic:stability_s} (b) $\emph{\textbf{F}}_{64,512}$ are examples of stable feature sets.

\subsection{A stability condition based on feature frequency}

In \cite{haug2020leveraging} the authors propose a metric for feature set stability based on statistical concepts. They measure the selection frequency of each feature and estimate its variance for a sequence of consecutive feature sets. They model the frequency of a feature using a Bernoulli process. The stability criterion we introduce requires that the variance becomes sufficiently small.

Given a feature set $\emph{\textbf{F}}_{k,t}$ and the original feature set $\emph{\textbf{F}}=\left \{ f_1,f_2,...,f_n \right \}$ ($k\in \mathbb{N},\ k\leq n$), we define the feature representation vector $\emph{\textbf{A}}_{k,t} \in \left \{ 0,1 \right \}^{n}$ as follows
\begin{equation}
    \emph{\textbf{A}}_{k,t}\left [ j \right ]=\left\{\begin{matrix}
1, & if\ f_j\in \emph{\textbf{F}}_{k,t}\\ 
0, & otherwise
\end{matrix}\right.
\end{equation}
We consider $r$ consecutive vectors and set the matrix  $\emph{\textbf{Z}}_{k,t,r}:=[\emph{\textbf{A}}_{k,t},...,\emph{\textbf{A}}_{k,t+r-1}]^{T}$. We define the stability of $\emph{\textbf{F}}_{k,t}$ for $r$ consecutive feature sets as:
\begin{equation}
\label{equ:sta}
    stab(\emph{\textbf{Z}}_{k,t,r}):=1-\frac{\frac{1}{n}\sum_{j=1}^{n}s_{j}^{2}}{\frac{k}{n}\left ( 1- \frac{k}{n}\right )}
\end{equation}
To understand the right side of equation (\ref{equ:sta}), we need to know that $p_{j}$ is the parameter of the Bernoulli process that models the occurrence of feature $f_j$. If we denote the empirical feature frequency with $\widehat{p_{j}}$, we can write $\widehat{p_{j}}=\frac{1}{r}\sum_{i=1}^{r}z_{ij}$ ($z_{ij}$ is an element of $\emph{\textbf{Z}}_{k,t,r}$). The empirical sample variance is given by $s_{j}^{2}=\frac{r}{r-1}\widehat{p_{j}}(1-\widehat{p_{j}})$. We know from theory that $s_{j}^{2}$ is an unbiased estimator of the variance of the feature frequency.

The stability metric $stab(\emph{\textbf{Z}}_{k,t,r})$ has values in $[-\frac{1}{r-1},1]$ \cite{nogueira2017stability}. If no feature shows any variance, i.e. each feature is either always selected or always left out, the value is 1. Otherwise, the value is less than 1. 

\begin{figure}[ht]
\centering
\subfigure[$stab(\emph{\textbf{Z}}_{k,t,r})$ vs $t$ in ARR]{
\begin{minipage}[t]{0.48\linewidth}
\centering
\includegraphics[width=1.7in]{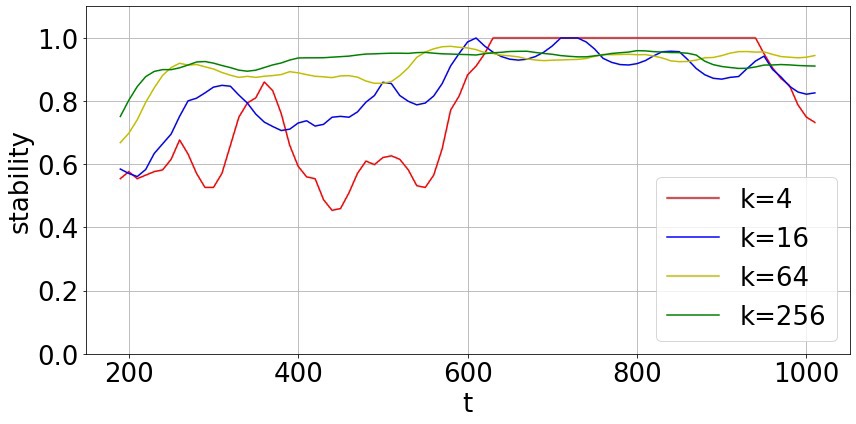}
\end{minipage}%
}%
\subfigure[$stab(\emph{\textbf{Z}}_{k,t,r})$ vs $t$ in LS]{
\begin{minipage}[t]{0.48\linewidth}
\centering
\includegraphics[width=1.7in]{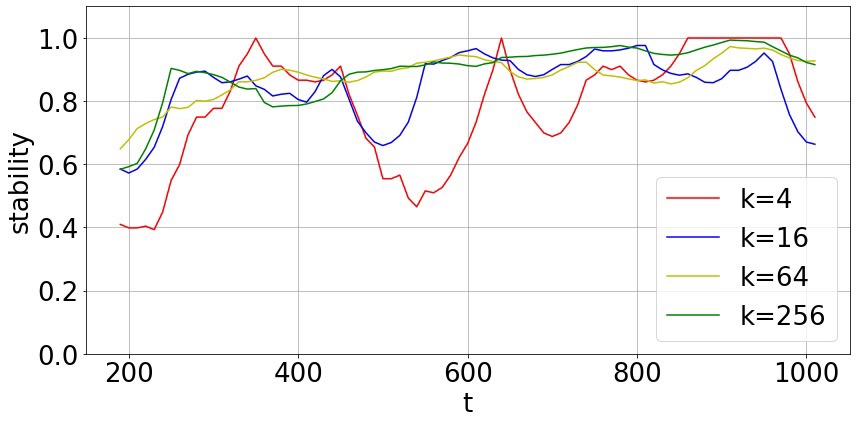}
\end{minipage}%
}%

\centering
\caption{The feature set stability of current feature sets over time using ARR and LS with samples from the KV flash-crowd data set and start point $t=1$.}
\label{pic:fea_stability}
\end{figure}

Similar to $sim(\emph{\textbf{F}}_{k, t_1},\emph{\textbf{F}}_{k, t_2})$ we observe using the data from our testbed that $stab(\emph{\textbf{Z}}_{k,t,r})$ tends to grow with the increase of $k$ and $t$. This effect becomes stronger the larger $k$ and $t$ are. Figure \ref{pic:fea_stability} shows an example with data from the KTH testbed running a KV service.  

We introduce the second heuristic notion of a \emph{stable feature set}: 
\begin{equation}
\label{equ:sta_con}
\begin{split}
    \emph{\textbf{F}}_{k,t}\ is\ stable\ for\ r,\ iff\ stab(\emph{\textbf{Z}}_{k,t,r}) > \eta \\
    for\ w\ consecutive\ timesteps
\end{split}
\end{equation}

Based on our experience with operational data, we set $r=10$, $w=10$ and $\eta=0.9$ for this work. This means that when 10 consecutive stability values are all larger than 0.9, then the feature set $\emph{\textbf{F}}_{k,t}$ is stable. In Figure \ref{pic:fea_stability} (a) $\emph{\textbf{F}}_{4,600}$ and in Figure \ref{pic:fea_stability} (b) $\emph{\textbf{F}}_{4,840}$ are examples of stable feature sets.

\subsection{Feature clustering}

We also investigated a different approach to determine whether a feature set $\emph{\textbf{F}}_{k,t}$ is stable. This approach includes clustering the features after $t$ samples have been obtained. The two conditions we formulated to identify a stable feature set were then applied on the feature clusters instead of on individual features. The motivation for this approach stems from the idea that features in the same cluster have the same information content for constructing the prediction model. Specifically, we have used DBSCAN \cite{scikitDBSCAN}, which is based on the cosine similarity metric to cluster the features. However, the evaluation showed that the accuracy of the obtained predictors has been only marginally better than when clustering was not used. For this reason, we have not included the detailed results in this paper.

\section{OSFS: online feature selection with low overhead}
\label{sec:online_feature_selection_with_low_overhead}

In this section, we introduce the Online Stable Feature Set algorithm (OSFS), which reads a stream of samples $\emph{\textbf{X}}_{1}$, $\emph{\textbf{X}}_{2}$, $\emph{\textbf{X}}_{3}$,... and returns the number of features $k$, the number of samples $t_{k}$ needed to determine $k$, and the stable feature set $\emph{\textbf{F}}_{k,t_k}$. 

Recall that the purpose of OSFS is to select a subset of available data sources (i.e. features), in order to reduce the monitoring costs and the training overhead for learning. To keep costs and overhead low, we want $k$ and $t_k$ to be small while still allowing for effective learning and prediction.

The standard use case for OSFS plays out as follows. We start monitoring the values of $n$ features at time $t=1$ and collect a sequence of samples with index $t=1,2,... $ We use the collected samples to find values for $k<<n$ and $t_{k}$ so that the first $t_{k}$ samples allow us to compute a stable feature set $\emph{\textbf{F}}_{k,t_k}$. At time  $t_{k}+1$, we reduce the number of data sources from $n$ to $k$ and continue motoring only sources from the set $\emph{\textbf{F}}_{k,t_k}$. 

\begin{figure}[ht]
 \centering
 \includegraphics[scale=0.4]{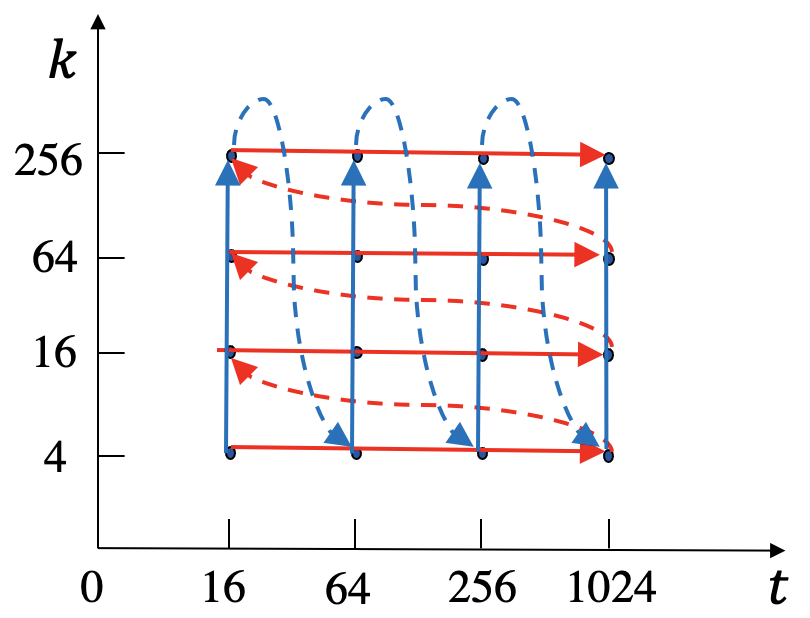}
 \caption{Search policy on one search space. The red path follows the policy $k-small$ and the blue path follows the policy $t-small$.}
 \label{fig:search_policy}
\end{figure}

We first present OSFS in a generic form, which allows for a range of instantiations and configurations. Algorithm \ref{OSFS} shows the options for configuration. First, we choose a feature ranking algorithm. Section \ref{sec:creating_ranked_feature_lists} contains two examples of such algorithms, namely, ARR and LS. Second, we choose a condition for a feature set  $\emph{\textbf{F}}_{k,t}$ to be stable. Section \ref{sec:Computing_stable_feature_sets} presents two possibilities of such conditions, namely, (\ref{equ:sim_con}) and (\ref{equ:sta_con}). Third, we choose the search space $S$, i.e. the space of possible values for $(k,t)$. An example of $S$ is a two-dimensional grid where $t$ has values $[16,64,256,1024]$ and $k$ has the values $[4,16,64,256]$. A search policy $P$ is an algorithm to traverse $S$. We have experimented with different search policies. For instance, policy $k-small$ starts with the smallest $k$, traverses all possible $t$ values, continues to the next larger $k$, traverses all possible $t$ values, etc. This way the grid is traversed from bottom to top (see Figure \ref{fig:search_policy} red path). Similarly, the policy $t-small$ starts with the smallest $t$, traverses all possible $k$ values, continues to the next larger $t$, traverses all possible $k$ values, etc. This way the grid is traversed from left to right (see Figure \ref{fig:search_policy} blue path). Many other policies are possible, for instance, random walk.

Once the execution of instantiated OSFS algorithm starts, initial values for $k$ and $t$ are selected (line 1). It then enters a loop where, during each iteration, one point in the grid is evaluated (line 2-7). First, the feature set $\emph{\textbf{F}}_{k,t}$ is computed using the ranking algorithm $A$ and available samples $\emph{\textbf{X}}_{1}$,..., $\emph{\textbf{X}}_{t}$. Then, the stability condition is evaluated for $\emph{\textbf{F}}_{k,t}$. If $\emph{\textbf{F}}_{k,t}$ is stable or the grid has been completely searched, the algorithm terminates and returns $\emph{\textbf{F}}_{k,t}$, $k$, $t$. Otherwise, the loop continues with the next $(k,t)$ pair.

When discussing the evaluations of OSFS in sections \ref{sec:evaluation_of_OSFS_on_different_data_sets} and \ref{sec:evaluation_of_OSFS_with_change_detection}, we present results obtained with feature ranking algorithms ARR and LS and with both feature set stability conditions. Regarding the search policy, most of the results in this paper are obtained with $k-small$, which gives priority to a feature set of small size at the possible expense of larger running time of OSFS. We also perform evaluations with policy $t-small$, which prioritises obtaining a stable feature set within a short amount of time at the possible expense of a large feature set.

\begin{algorithm} [ht]
\label{OSFS}
  \caption{Online Stable Feature Set (OSFS)}
  \KwIn{Sample sequence $\emph{\textbf{X}}_{1}$, $\emph{\textbf{X}}_{2}$, $\emph{\textbf{X}}_{3}$,...;\\
  \ \ \ \ \ \ \ \ \ Feature ranking algorithm $A$;\\ 
  \ \ \ \ \ \ \ \ \ Metric and condition for stable feature set $C$;\\
  \ \ \ \ \ \ \ \ \ Search space $S$ for $(k,t)$;\\
  \ \ \ \ \ \ \ \ \ Search policy $P$ to traverse $S$.}  
  \KwOut{Stable feature subset $\emph{\textbf{F}}_{k,t}$, $k$, $t$.\\
  }
 Initialize $(k,t)$\;
 \While{True}{
    Compute $\emph{\textbf{F}}_{k,t}$ using $A$ on $\emph{\textbf{X}}_{1}$,..., $\emph{\textbf{X}}_{t}$\;
    \If{$\emph{\textbf{F}}_{k,t}$ is stable or $S$ is exhausted}
    {
        return $\emph{\textbf{F}}_{k,t}$, $k$, $t$\;
    }
    \Else{
        choose next $(k,t)$ following $P$\;
    }
    }
\end{algorithm}

\begin{algorithm} [ht]
\label{OSFS-sim}
  \caption{OSFS-ARR-$sim$-$k-small$}  
  \KwIn{Sample sequence $\emph{\textbf{X}}_{1}$, $\emph{\textbf{X}}_{2}$, $\emph{\textbf{X}}_{3}$,...;\\
  \ \ \ \ \ \ \ \ \ Feature ranking algorithm ARR;\\ 
  \ \ \ \ \ \ \ \ \ Metric $sim$ and condition (\ref{equ:sim_con}) for stable feature \\
  \ \ \ \ \ \ \ \ \ set;\\
  \ \ \ \ \ \ \ \ \ Search space $[4-256]*[8-1024]$ for $(k,t)$;\\
  \ \ \ \ \ \ \ \ \ Search policy $k-small$.
  }  
  \KwOut{Feature subset $\emph{\textbf{F}}_{k,t_k}$, $k$, $t_{k}$\\
  $subset(k,t,ARR)$ returns top $k$ features computed with ARR using samples with index $1,...,t$. }
 $\eta=0.5$ (threshold for stable feature set)\;
 $read=false$\;
 \For{$k$ in [4,16,64,256]}{
    \If{$not \ read$}{
    read and store $\emph{\textbf{X}}_{t}, t=1,...,16$
    }
  $\emph{\textbf{F}}_{k1}=subset(k,8,ARR)$\;
  $\emph{\textbf{F}}_{k2}=subset(k,16,ARR)$\;
  $sim_{k12}=sim(\emph{\textbf{F}}_{k1},\emph{\textbf{F}}_{k2})$;

  \For{$t= 17, ..., 1024$}
  {
    \If{$not \ read$}
    {
        read and store $\emph{\textbf{X}}_{t}$
   }
       \If{$t$ in [32,64,128,256,512,1024]}
    {

  $\emph{\textbf{F}}_{kt}=subset(k,t,ARR)$\;
    $sim_{kt}=sim(\emph{\textbf{F}}_{k2},\emph{\textbf{F}}_{kt})$\;
    \If{$sim_{kt}<sim_{k12}\ and\ sim_{k12}>\eta$}
    {
        return $\emph{\textbf{F}}_{k1},k,t/4$\;
    }
    \ElseIf{$sim_{kt}>\eta\ and\ t==1024$}{
    return $\emph{\textbf{F}}_{k2},k,t/2$\;
    }
    \Else{
    $\emph{\textbf{F}}_{k1}=\emph{\textbf{F}}_{k2}$\;
    $\emph{\textbf{F}}_{k2}=\emph{\textbf{F}}_{kt}$\;
    $sim_{k12}=sim_{kt}$;
    }
    }
    }
    $read=true$
       }
  return $\emph{\textbf{F}}_{k2},256,1024$  
\end{algorithm} 

Algorithm \ref{OSFS-sim} is an example of an instantiated OSFS algorithm, which we use for evaluation. The algorithm takes as input the sequence of arriving samples $\emph{\textbf{X}}_{t}$ and is initialized with the feature ranking algorithm ARR. ARR is used in the function $subset()$, which takes as input $k$ and the first $t$ samples and returns a set with the top $k$ features ranked by ARR.

It has two main loops: an outer loop (lines 3-23) that iterates over a subspace of $k$ and an inner loop (lines 9-22) that iterates over a subspace of $t$. The index values of $k$ and $t$ increase exponentially to enable the exploration of a large space with a small number of evaluations. The algorithm performs a grid search in the space of tuples $(k,t)$ with the termination condition A: $sim_{kt}<sim_{k12}\ and\ sim_{k12}>\eta$ (line 15) or B: $sim_{kt}>\eta\ and\ t==1024$ (line 17). In case the conditions A or B are never met, the algorithm terminates after the search on the grid sector $[4-256]*[8-1024]$ has been completed. The key termination condition A expresses the case where (1) the similarity of two consecutive feature sets is above the threshold $\eta$ and (2) the similarity declines when the subsequent feature set is considered. 

The algorithm reads at least 32 samples in order to ensure statistical viability and terminates after at most 1024 samples. Since the outer loop is indexed by increasing $k$, it favors a smaller $k$ at the possible expense of a larger $t$. This means that we prefer reducing the monitoring overhead over reducing the time to compute $\emph{\textbf{F}}_{k,t}$.

\section{Recomputing feature set after concept drift}
\label{sec:change_detection_and_model_recomputation}
The distribution from which the samples are drawn for training prediction models can evolve over time, and this phenomenon is referred to as \emph{concept drift}, which is a well-studied topic in data analysis \cite{gama2014survey}. When concept drift occurs, the prediction model must generally be recomputed in order to maintain prediction accuracy. In addition, the feature set from which the prediction model is computed should be adapted as well for the same reason.

In the context of network systems and their operation, concept drift can occur when a system is reconfigured or when the resource allocation policy for services or applications is changed. This means that, when concept drift occurs, OSFS must be run again to identify the updated set of features for recomputing the prediction model.

In this work, we use a state-of-the-art concept drift detection algorithm called STUDD (Student-Teacher Method for Unsupervised Concept Drift Detection) \cite{cerqueira2021studd} on our data traces. STUDD detects concept drift in an unsupervised manner by using a student-teacher paradigm \cite{bucilu2006model}. The basic idea of this paradigm is that in addition to the primary predictor, which is called \emph{teacher}, a second predictor, which is called \emph{student}, is trained. The student model is learned on the same input dataset \emph{\textbf{X}} as the teacher, but the target set \emph{\textbf{Y}} is replaced with the values predicted by the teacher. During the prediction phase, the discrepancy between the teacher's prediction and the student's prediction for the same target values is monitored. The Page-Hinkley test \cite{page1954continuous} is applied to detect changes in the time series of the discrepancy values. Such a change indicates a concept drift event.

We use a random forest regressor to compute the teacher as well as the student models.

In Section \ref{sec:evaluation_of_OSFS_with_change_detection} we study the effect of recomputing the stable feature set on the prediction accuracy using the KTH dataset. Specifically, we compare the accuracy values for scenarios where the stable feature set is recomputed after change detection, with scenarios where the stable feature set is computed only at the beginning of the trace.

\section{Traces for evaluation}
\label{sec:testbed}

\subsection{KTH testbed traces}
\begin{figure}[ht]
 \centering
 \includegraphics[scale=0.23]{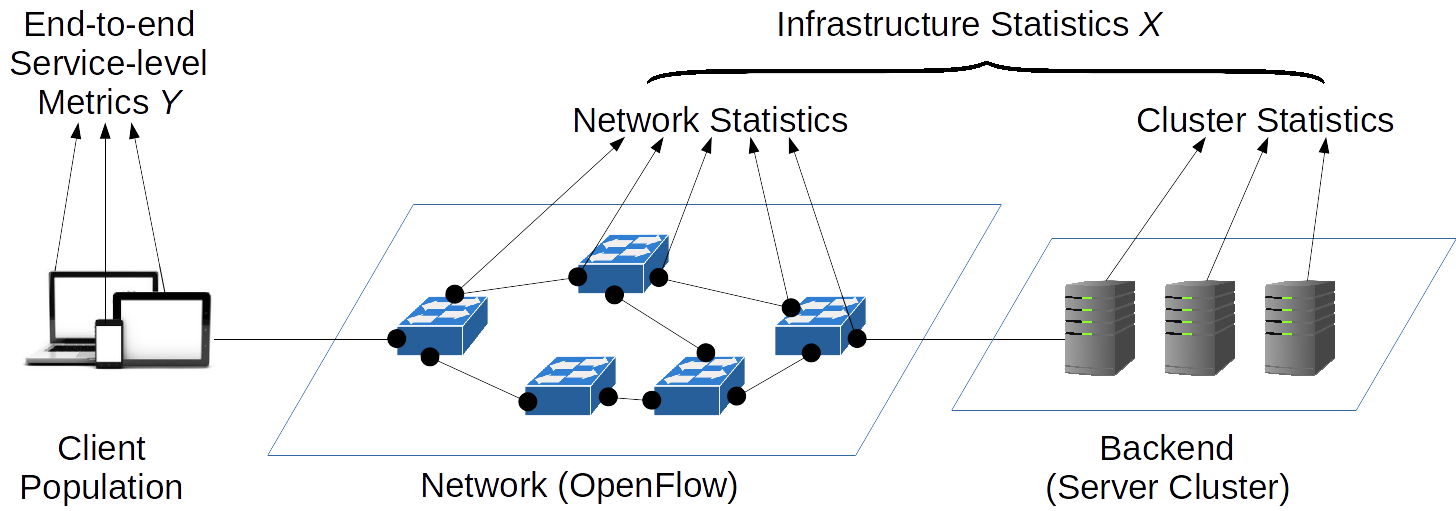}
 \caption{The testbed at KTH, providing the infrastructure for experiments. In various scenarios we predict end-to-end service-level metrics from low-level infrastructure measurements \cite{stadler2017learning}.}
 \label{fig:testbed}
\end{figure}

Figure \ref{fig:testbed} outlines our laboratory testbed at KTH. It includes a server cluster, an emulated OpenFlow network, and a set of clients. A more detailed description of the KTH testbed setup and the services running on it is given in \cite{stadler2017learning}. 

The online feature selection method proposed in this paper has been evaluated using data from four experiments. Two of them involve running a Video-on-Demand (VoD) service and two a Key Value store (KV) service.

\begin{enumerate}
\item \textit{VoD periodic}: In this experiment, we run the VoD service under a periodic load pattern on the testbed. Data is collected every second over a period of $50\,000$ seconds. After cleaning the dataset, it contains $50\,000$ samples and $1\,296$ features. More details about the experiment are given in \cite{yanggratoke2015predicting}.

\item \textit{VoD flash-crowd}: This experiment relies on the same setup as VoD periodic, except that the testbed is loaded with a flash-crowd pattern. After cleaning the dataset, $50\,000$ samples and $1\,255$ features remain. More details about the experiment are given in \cite{yanggratoke2015predicting}. 

\item \textit{KV periodic}:
In this experiment, we run the KV service under a periodic load pattern on the testbed. Measurements are collected every second over a period of $28\,962$ seconds. The dataset contains $10\,374$ features collected from the server cluster and $176$ features from the network switches. After cleaning the data set, $1\,751$ features remain. More details about the experiment are given in \cite{stadler2017learning}.

\item \textit{KV flash-crowd}:
This experiment relies on the same setup as KV periodic, except that the testbed is loaded with a flash-crowd pattern. After cleaning the dataset, $19\,444$ samples and $1\,723$ features remain. More details about the experiment are given in \cite{stadler2017learning}.

\end{enumerate}

Table \ref{tab:kth_dataset} summarizes the basic information of these four traces. All four traces are available at \cite{CNSM2019-traces}.

\begin{table}[H]
    \centering
    \caption{Datasets from KTH test bed}
    \label{tab:kth_dataset}
    \setlength{\tabcolsep}{0.8mm}{
    \begin{tabular}{|c|c|c|c|}
    \hline
    Dataset&Number of features&Number of samples&Target\\
    \hline
    KV flash crowd&1723&19444&$'ReadsAvg'$\\
    \hline
    KV periodic&1751&28962&$'ReadsAvg'$\\
    \hline
    VoD flash crowd&1255&50000&$'DispFrames'$\\
    \hline
    VoD periodic&1296&50000&$'DispFrames'$\\
    \hline
    \end{tabular}}
\end{table}

\subsection{FedCSIS 2020 Challenge dataset}

The FedCSIS 2020 Challenge dataset \cite{janusz2020network} is a public dataset provided by EMCA Software, an analytics company. It contains around 2000 samples collected from December 2019 to February 2020. Each sample aggregates one hour of measurements, and there are 24 samples per day. The data are gathered from 3728 hosts. Hosts can have different numbers of attributes. For each attribute, such as CPU utilization, several feature columns are included, for example, the mean value, the standard deviation. From the challenge dataset, we construct two smaller datasets for evaluation. In each of these two datasets, all samples share the same features.

\begin{itemize}
    \item Dataset1 contains samples from hosts that have the following six attributes $['cpu\_1m','cpu\_5m','cpu\_5s','memoory\_free',$ $'memory\_total','memory\_used']$. For every attribute, four feature columns are included with the mean, the standard deviation, the minimum and the maximum value, respectively.
    \item Dataset2 contains samples from hosts that have the following six attributes $['cpuusagebyproc','memoryallocatedbyproc',$ $'in\_traffic','out\_traffic','error\_in','error\_out']$. For every attribute, four feature columns are included with the mean, the standard deviation, the minimum and the maximum value, respectively.
\end{itemize}

In these two datasets, many feature values are missing. We remove all feature columns where more than 30\% of the values are missing. The remaining missing values are estimated through linear interpolation. After these steps, the number of features and the number of samples in each dataset are listed in Table \ref{tab:dataset}.

\begin{table}[H]
    \centering
    \caption{Datasets from FedCSIS 2020 Challenge dataset}
    \label{tab:dataset}
    \begin{tabular}{|c|c|c|}
    \hline
    Dataset&Number of features&Number of samples\\
    \hline
    Dataset1&30438&1917\\
    \hline
    Dataset2&10163&1917\\
    \hline
    \end{tabular}
\end{table}

\section{Evaluation of OSFS on different data sets}
\label{sec:evaluation_of_OSFS_on_different_data_sets}

We evaluate OSFS by asking two main questions: 
\begin{enumerate}
    \item How effective is OSFS in reducing communication and computing overhead as well as the data collection time, i.e. the number of samples needed for feature selection? What are typical values for $k$ and $t_k$ for the KTH and FedCSIS datasets?
    \item How effective is OSFS in producing feature sets for training accurate prediction models?
\end{enumerate}

To obtain the evaluation results, we perform two preprocessing steps on the datasets. First, we linearly scale the values of each feature vector to the range $[0,1]$. Second, we remove the feature vectors with a variance below 0.0001. For prediction, we use a random forest regressor from the \emph{scikit-learn} library \cite{scikitrandom}. As for hyperparameters, we use 100 trees and the default values for the other parameters. The prediction error is expressed as Normalized Mean Absolute Error (NMAE), which is computed as follows:
\begin{equation}
\label{equ:NMAE}
  NMAE(\widehat{y_{1}},...,\widehat{y_{q}})=\frac{1}{\overline{y}}(\frac{1}{q}\sum_{i=1}^{q}\left | y_{i}-\widehat{y_{i}} \right |)  
\end{equation} 
where $\overline{y}$ is the mean value of targets in the test set,  $q$ is the number of samples and $\widehat{y_{i}}$ is the predicted value.

Table \ref{tab:kth_res} shows the evaluations for the KTH datasets and Table \ref{tab:challenge_res} for the FedCSIS datasets. Both tables have the same column structure. The first column identifies the dataset, the second column lists the feature ranking algorithm, the third column gives the feature set stability metric, the fourth column gives the search policy. The fifth and sixth columns give the output of OSFS, the seventh and eighth columns provide the prediction error. 

Each row in Tables \ref{tab:kth_res} and \ref{tab:challenge_res} shows the result from an evaluation scenario for a given dataset, feature ranking algorithm, feature set stability metric and search policy. A scenario contains ten evaluation runs of OSFS with different start points on the dataset (whereby one start point is t=1 and the other 9 are chosen uniformly at random between t=2 and t=10\ 000). 

We observe that OSFS produces a range of feature set sizes $k$, from 4 to 212. The same applies to the number of samples $t_k$ (i.e. the data collection time) with a range from 8 to 882. As expected, the search policy $k-small$ produces small values for $k$ and larger values for $t_k$, and for the search policy $t-small$, the opposite applies. In all scenarios, we see a significant reduction of the size of the feature sets compared with the total number of features. We also find that, while the values for $k$ and $t_k$ are strongly dependent on the search policy, they are less influenced by other parameters, namely, the type of dataset and the feature ranking algorithm. 

The column \textbf{Online FS error} shows the error of a prediction model trained with the feature set of size $k$ that has been produced with $t_k$ samples. These values reflect the capability of OSFS to produce an effective feature set for model training. We compare the results with two baseline methods. The first baseline method uses all samples of the dataset for feature selection. The values for this baseline are listed in column \textbf{Offline FS error}. The second baseline method does not rely on feature selection and uses all features to train the prediction model. The result for this method is listed in the lowest row of each dataset and indicated by \textbf{No FS}. 

Comparing the values in column Offline FS error with the error values in No FS shows us the cost of reducing the feature set. Depending on the scenario, the difference in error ranges from 2\% to 115\% for the KTH datasets and from 158\% to 369\% for the FedCSIS datasets. (We explain the larger difference on the FedCSIS datasets with the fact that the complete feature space of the FedCSIS datasets is between 5 and 16 times larger than that of the KTH datasets, while the number of selected features by OSFS is almost the same.) While the difference in the error can be quite significant, we have shown in earlier research that it can be effectively reduced by using a supervised feature ranking algorithm instead of an unsupervised one \cite{samani2019efficient}. 

Most importantly, the values in the column Online FS error tend to be somewhat smaller than those in Offline FS error, 6\% on KTH datasets and 25\% on FedCSIS datasets in average. This result suggests that OSFS is effective as an online feature selection algorithm and is robust regarding the sample interval on the trace that is used for feature selection.

\begin{table*}[ht]
    \centering
    \caption{Evaluation of OSFS on KTH data sets. The prediction method is random forest. The error is expressed in NMAE (\ref{equ:NMAE}). Each row shows the result from an evaluation scenario with ten runs.}
    \label{tab:kth_res}
    \begin{tabular}{|c|c|c|c|c|c|c|c|}
    \hline
     Dataset&Method&Metric&Search&$k$ &$t_k$& \textbf{Online FS error}&Offline FS error\\
    \hline
    KV flash-crowd&ARR&Similarity&$k-small$&$25.6\pm 25.6$&$251.2\pm 224.0$&$\textbf{0.0280}\pm 0.0056$&$0.0304\pm 0.0049$\\
    \cline{3-8}
    ~&~&Stability&$k-small$&$17.2\pm 16.5$&$744\pm 159.6$&$\textbf{0.0299}\pm 0.0054$&$0.0322\pm 0.0033$\\
    \cline{4-8}
    ~&~&~&$t-small$&$192.4\pm 98.4$&$472\pm 56.9$&$\textbf{0.0233}\pm 0.0060$&$0.0227\pm 0.0043$\\
    \cline{2-8}
    ~&LS&Similarity&$k-small$&$10\pm 6$&$64\pm 72.3$&$\textbf{0.0232}\pm 0.0014$&$0.0255\pm 0.0007$\\
     \cline{3-8}
    ~&~&Stability&$k-small$&$6.4\pm 4.8$&$724\pm 244.7$&$\textbf{0.0289}\pm 0.0054$&$0.0322\pm 0.0037$\\
    \cline{4-8}
    ~&~&~&$t-small$&$186.4\pm 107.4$&$472\pm 83$&$\textbf{0.0235}\pm 0.0060$&$0.0234\pm 0.0054$\\
    \cline{2-8}
    ~&No FS&~&~&1723&~&~&0.0184\\
    \hline
    KV periodic&ARR&Similarity&$k-small$&$10\pm 6$&$249.6\pm 194.4$&$\textbf{0.0325}\pm 0.0071$&$0.0434\pm 0.0006$\\
    \cline{3-8}
    ~&~&Stability&$k-small$&$8.8\pm 5.9$&$660\pm 179.5$&$\textbf{0.0294}\pm 0.0015$&$0.0435\pm 0.0006$\\
    \cline{4-8}
    ~&~&~&$t-small$&$211.6\pm 89.8$&$415\pm 50.2$&$\textbf{0.0253}\pm 0.0011$&$0.0263\pm 0.0059$\\
    \cline{2-8}
    ~&LS&Similarity&$k-small$&$12.4\pm 5.5$&$195.2\pm 177.3$&$\textbf{0.0268}\pm 0.0016$&$0.0264\pm 0.0001$\\
     \cline{3-8}
    ~&~&Stability&$k-small$&$14.8\pm 17.3$&$514\pm 180.5$&$\textbf{0.0273}\pm 0.0014$&$0.0262\pm 0.0005$\\
    \cline{4-8}
    ~&~&~&$t-small$&$122.8\pm 111.2$&$451\pm 68.3$&$\textbf{0.0245}\pm 0.0025$&$0.0249\pm 0.0010$\\
    \cline{2-8}
    ~&No FS&~&~&1751&~&~&0.0214\\
    \hline
    VoD flash-crowd&ARR&Similarity&$k-small$&$12.4\pm 5.5$&$131.2\pm 142.7$&$\textbf{0.1629}\pm 0.0139$&$0.1328\pm 0.0083$\\
    \cline{3-8}
    ~&~&Stability&$k-small$&$16\pm 17.0$&$724\pm 198.1$&$\textbf{0.1340}\pm 0.0067$&$0.1349\pm 0.0087$\\
    \cline{4-8}
    ~&~&~&$t-small$&$140.8\pm 94.1$&$387\pm 74.6$&$\textbf{0.1081}\pm 0.0255$&$0.1129\pm 0.0222$\\
    \cline{2-8}
    ~&LS&Similarity&$k-small$&$17.2\pm 16.5$&$226.4\pm 198.2$&$\textbf{0.0969}\pm 0.0405$&$0.1469\pm 0.0257$\\
     \cline{3-8}
    ~&~&Stability&$k-small$&$7.6\pm 5.5$&$749\pm 213.2$&$\textbf{0.1170}\pm 0.0338$&$0.1657\pm 0.0135$\\
    \cline{4-8}
    ~&~&~&$t-small$&$167.2\pm 110.4$&$520\pm 85.2$&$\textbf{0.0817}\pm 0.0320$&$0.0899\pm 0.0425$\\
    \cline{2-8}
    ~&No FS&~&~&1255&~&~&0.0771\\
    \hline
    VoD periodic&ARR&Similarity&$k-small$&$10\pm 6$&$363.2\pm 196.5$&$\textbf{0.2085}\pm 0.0203$&$0.1795\pm 0.0041$\\
    \cline{3-8}
    ~&~&Stability&$k-small$&$5.2\pm 3.6$&$781\pm 168.2$&$\textbf{0.2123}\pm 0.0189$&$0.1827\pm 0.0024$\\
    \cline{4-8}
    ~&~&~&$t-small$&$174.4\pm 100.8$&$356\pm 59.2$&$\textbf{0.1413}\pm 0.0196$&$0.1384\pm 0.0259$\\
    \cline{2-8}
    ~&LS&Similarity&$k-small$&$25.6\pm 25.6$&$108\pm 85.5$&$\textbf{0.1162}\pm 0.0295$&$0.1426\pm 0.0718$\\
     \cline{3-8}
    ~&~&Stability&$k-small$&$17.2\pm 16.5$&$686\pm 145.9$&$\textbf{0.1139}\pm 0.0605$&$0.1166\pm 0.0739$\\
    \cline{4-8}
    ~&~&~&$t-small$&$211.6\pm 89.8$&$541\pm 115.7$&$\textbf{0.1130}\pm 0.0084$&$0.1243\pm 0.0344$\\
    \cline{2-8}
    ~&No FS&~&~&1296&~&~&0.1190\\
    \hline
\end{tabular}
\end{table*}

\begin{table*}[ht]
    \centering
    \caption{Evaluation of OSFS on FedCSIS 2020 Challenge data sets. The prediction method is random forest. The error is expressed in NMAE (\ref{equ:NMAE}). Each row shows the result from an evaluation scenario with ten runs. For dataset1, the prediction target is $'host4846\_cpuusagebyproc\_mean'$ and for dataset2 is $'host0958\_cpu\_1m\_mean'$.}
    \label{tab:challenge_res}
    \begin{tabular}{|c|c|c|c|c|c|c|c|}
    \hline
     Dataset&Method&Metric&Search&$k$ &$t_k$& \textbf{Online FS error}&Offline FS error\\
    \hline
    Dataset1&ARR&Similarity&$k-small$&$26.8\pm 24.9$&$217.6\pm 169.2$&$\textbf{0.701}\pm 0.124$&$0.593\pm 0.243$\\
    \cline{3-8}
    ~&~&Stability&$k-small$&$10\pm 6$&$600\pm 135.2$&$\textbf{0.829}\pm 0.132$&$0.730\pm 0.218$\\
    \cline{2-8}
    ~&LS&Similarity&$k-small$&$10\pm 18$&$8\pm 0$&$\textbf{0.291}\pm 0.024$&$0.884\pm 0.170$\\
     \cline{3-8}
    ~&~&Stability&$k-small$&$44.8\pm 73.8$&$876\pm 66.7$&$\textbf{0.569}\pm 0.197$&$0.715\pm 0.249$\\
    \cline{2-8}
    ~&No FS&~&~&10162&~&~&0.230\\
    \hline
    Dataset2&ARR&Similarity&$k-small$&$168.4\pm 108.7$&$153.6\pm 145.1$&$\textbf{0.221}\pm 0.128$&$0.208\pm 0.103$\\
    \cline{3-8}
    ~&~&Stability&$k-small$&$42.4\pm 26.7$&$882\pm 65.2$&$\textbf{0.275}\pm 0.121$&$0.272\pm 0.116$\\
    \cline{2-8}
    ~&LS&Similarity&$k-small$&$4\pm 0$&$92.8\pm 36.3$&$\textbf{0.055}\pm 0.004$&$0.218\pm 0$\\
     \cline{3-8}
    ~&~&Stability&$k-small$&$4\pm 0$&$337\pm 69.7$&$\textbf{0.055}\pm 0.003$&$0.218\pm 0$\\
    \cline{2-8}
    ~&No FS&~&~&30437&~&~&0.058\\
    \hline
\end{tabular}
\end{table*}

Other conclusions we draw from Tables \ref{tab:kth_res} and \ref{tab:challenge_res}:

\begin{itemize}
    \item OSFS can achieve a massive reduction in the size of the feature set,  1-3 orders of magnitude. 
    \item The number of samples needed to compute the feature sets averages around 430 on the investigated datasets. For the KTH testbed, where metrics are monitored once per second, this means that OSFS produces a stable feature set within an average of 7 minutes. 
    \item The feature ranking algorithm LS generally provides much better results than ARR, although at the cost of higher computational complexity when the number of samples used for feature selection becomes large (see Section \ref{sec:creating_ranked_feature_lists}). 
    \item Consistent with our earlier results (e.g., \cite{im2015_realm, stadler2017learning}), we find that the type of service and the load pattern significantly affect the prediction error.
    \item We find that neither the feature set similarity metric nor the feature set stability metric outperforms the other on the datasets we investigated. Since the feature set similarity has a lower computational complexity, we prefer this metric for future investigations.
    \item On the investigated datasets, OSFS achieves the best results, on average, with feature ranking algorithm LS, search policy $k-small$, and the feature set similarity metric.
\end{itemize}

\section{Evaluation of OSFS with online training and change detection}
\label{sec:evaluation_of_OSFS_with_change_detection}

In Section \ref{sec:evaluation_of_OSFS_on_different_data_sets}, we evaluate OSFS by studying the prediction error of a random forest regressor. In this case, feature selection is performed online, but model training of the regressor is performed offline. The column Online FS error in Tables \ref{tab:kth_res} and \ref{tab:challenge_res} gives the evaluation results. 

Building on these results, we investigate two key questions in this section. First, which prediction accuracy can be achieved when not only feature selection is performed online, but also model training, and how do the results compare to offline training? Second, can better prediction accuracy be achieved if the feature set is adapted over time? We address the second question through detecting concept drift, i.e. a change in the conditional distribution $P(Y|X)$ over time (see Section \ref{sec:related_work}). We use the STUDD algorithm for the concept drift detection (see Section \ref{sec:change_detection_and_model_recomputation}).

The implementation of the Page-Hinkley test in STUDD uses the \emph{scikit-multiflow} library \cite{montiel2018scikit} with the change parameter $delta$ set to 0.05.  Regarding OSFS, we use the $k-small$ search policy and the $similarity$ metric.

Table \ref{tab:OSFS_change} summarizes the results of these evaluations. The first column indicates the dataset, the second shows the feature ranking algorithm, the third column (\textbf{Offline train}) gives the prediction error when using online feature selection and offline training. (The values are computed the same way as in Table \ref{tab:kth_res} column Online FS error.) The fourth column (\textbf{Online train}) shows the prediction error when using online feature selection and online training. The fifth column (\textbf{Model retrain}) gives the results when using online feature selection, online training and model re-training after concept drift is detected. The sixth column (\textbf{Model retrain and feature recomp}) provides the most important results. It shows the prediction error when using online feature selection, online learning as well as feature set re-computation and model re-training after detecting concept drift. Finally, the last column (change) indicates the number of changes in concepts discovered on the trace. 

\begin{figure*}[ht]
 \centering
 \includegraphics[scale=0.6]{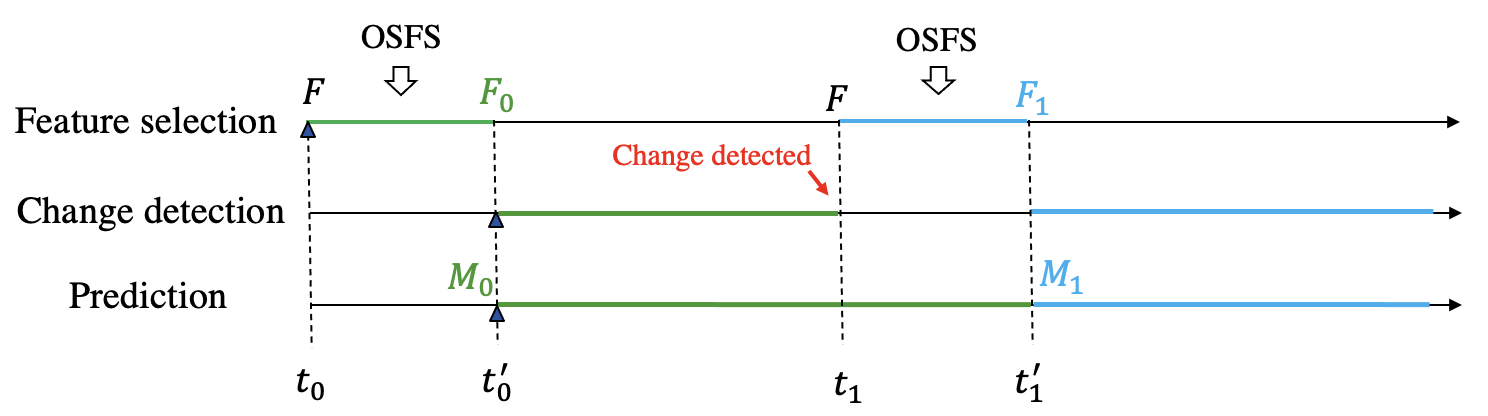}
 \caption{OSFS selects subset $F_0$ from the entire feature set $F$ using samples from interval [$t_0$,$t_{0}^{'}$]. When a change is detected at time $t_1$, OSFS selects subset $F_1$ from $F$ using samples from interval [$t_1$,$t_{1}^{'}$].}
 \label{fig:error}
\end{figure*}

Each row in Table \ref{tab:OSFS_change} lists the result from a run with start point $t=1$. Figure \ref{fig:error} describes how an experiment takes place. At the start point $t_0$, the system begins with periodically collecting samples with values from the entire feature set $\emph{\textbf{F}}$. At time $t_{0}^{'}$, OSFS returns a stable feature set $\emph{\textbf{F}}_{0}$ (based on the samples in the interval [$t_0$,$t_{0}^{'}$]). At this point, the monitoring system begins to collect samples with values only from the reduced feature set $\emph{\textbf{F}}_{0}$. Based on this feature set, the prediction model $M_0$ is trained and the system starts predicting target values. In the case of online training, the samples in the interval [$t_0$,$t_{0}^{'}$] are used. At time $t_1$, the change detection algorithm detects a concept drift and triggers OSFS to recompute the feature set. At this point, the monitoring system switches to monitor the values from the entire feature set $\emph{\textbf{F}}$. The newly computed feature set $\emph{\textbf{F}}_{1}$ becomes available at time $t_{1}^{'}$. In the case of model re-training, the new predictor $M_1$ is trained with the samples from the interval [$t_1$,$t_{1}^{'}$]. For reason of simplicity, the training time is omitted in Figure \ref{fig:error}.

From Table \ref{tab:OSFS_change}, we gain the following insights regarding the performance of OSFS on the studied datasets:
\begin{itemize}
    \item Feature set re-computation is effective as it can achieve a reduction in error by 3\% to 6\%, in addition to the reduction achieved by model re-training (8\% to 80\%). 
    \item Online training (Online train), compared with offline training (Offline train), incurs 18\% to 955\% higher prediction error. This error can be reduced by considering additional samples for training. Note that decreasing the difference in prediction error between online training and offline training is not studied in this work, as the focus is on online feature selection.
\end{itemize}
In summary, the error of a predictor can be significantly reduced by recomputing the feature set and retraining the model after a concept drift.

\begin{table*}[ht]
    \centering
    \caption{Evaluation of OSFS for online model training and model re-training under change detection. The prediction method is random forest. The error is expressed in NMAE (\ref{equ:NMAE}). The most important column is 'Model retrain and feature recomp', which shows he  prediction  error  when  using online  feature  selection,  online  learning  as  well  as  feature set re-compuation and  model re-training after detecting concept drift. }
    \label{tab:OSFS_change}
    \begin{tabular}{|c|c|c|c|c|c|c|}
    \hline
    Dataset&Method&Offline\ train&Online\ train&Model\ retrain&Model\ retrain\ and\ feature\ recomp&change\\
    \hline
    KV flash-crowd&ARR&0.0238&0.0387&0.0387&0.0387&0\\
    \cline{2-7}
    ~&LS&0.0232&0.2447&0.0389&\textbf{0.0393}&1\\
    \hline
    KV periodic&ARR&0.0436&0.1554&0.0579&\textbf{0.0531}&2\\
    \cline{2-7}
    ~&LS&0.0261&0.0985&0.0588&\textbf{0.0534}&2\\
    \hline
    VoD flash-crowd&ARR&0.1499&0.1163&0.1163&0.1163&0\\
    \cline{2-7}
    ~&LS&0.0654&0.1157&0.1157&0.1157&0\\
    \hline
    VoD periodic&ARR&0.2280&0.2700&0.1789&\textbf{0.1789}&1\\
    \cline{2-7}
    ~&LS&0.1844&0.1840&0.1687&\textbf{0.1684}&1\\
    \hline
    \end{tabular}
\end{table*}

\section{Related work}
\label{sec:related_work}

In the context of this paper, we understand \emph{online learning} as an iterative learning method where training samples arrive sequentially as a data stream. Periodic monitoring of infrastructure metrics in a networked system creates such a stream of samples. The learning algorithms that enable learning on a stream of samples are often referred to as \emph{incremental learning}. \cite{gepperth2016incremental} formalizes the concept of incremental learning and discusses algorithms that require only constant memory. In \cite{losing2018incremental}, the authors compare eight incremental learning algorithms for supervised classification. Instead of the term incremental learning, many authors use the expression \emph{learning on data streams}. \cite{bifet2018machine} is a textbook that provides a state-of-the-art description of this subject. Algorithms for data stream mining including classification, regression, clustering, and frequent pattern mining are discussed. Most of these algorithms are adapted from well-known offline learning algorithms. 

Most works that address learning on data streams assume that the samples in a data stream are drawn from a distribution $P(X,Y)$, which is also referred to as a concept \cite{haug2021learning}. The change of such a concept is called a \emph{concept drift}. For our research, it is important to detect a change in the conditional distribution $P(Y|X)$, which is called a \emph{real concept drift} \cite{haug2021learning}. In this paper, we use the STUDD algorithm \cite{cerqueira2021studd} for detecting a change in the conditional probability distribution $P(Y|X)$ (see Section \ref{sec:change_detection_and_model_recomputation}). \cite{gama2014survey} provides a survey on concept drift adaptation methods, which detect concept drift and adapt the prediction model. The paper contains a taxonomy of such methods. It also lists metrics for the evaluation of change detection methods.

A key challenge in learning on operational data from networked systems, which are distributed by nature, is respecting the resource constraints. The extraction of data on the devices, their transportation, the data preprocessing, and the model training require network, computing and memory resources. Much research has been conducted to reduce resource consumption while still being able to produce accurate prediction models. One research direction is \emph{online sample selection} where samples are kept in a cache of constant size for later processing and model training. A sample selection algorithm decides which samples should be stored in the cache and which should be dropped from the data stream. A well-known and efficient algorithm is reservoir sampling, which chooses a sample set of fixed size uniformly at random from a population of unknown size in a single pass \cite{vitter1985random}. In earlier work \cite{villacca2021online} we have evaluated four online sampling algorithms which have been derived from known algorithms. We specifically argue that feature selection algorithms can be adapted for sample selection and provide experimental evidence that the resulting algorithms can be effective.

A second research direction focuses on reducing the dimensionality of the feature space, which corresponds to the number of features. Many methods have been developed that map a high-dimensional feature space into a low-dimensional space. Principal Components Analysis (PCA) \cite{goodfellow2016deep} is a linear and popular method that is often used for this purpose. A second method, which is based on neural networks, is auto-encoder \cite{goodfellow2016deep}. Another approach to reduce the dimensionality of the feature space is feature selection. In this paper, we study feature selection methods, because they allow us to significantly reduce monitoring costs, in addition to saving model training costs. The other dimensionality reduction methods described above only reduce model training costs.  

Feature selection has been studied as an effective data preprocessing strategy in both the machine learning and data mining fields for several decades. Many survey papers cover and compare feature selection methods, e.g. \cite{guyon2003introduction}\cite{chandrashekar2014survey}\cite{li2017feature}\cite{solorio2020review}\cite{di2021supervised}. For instance, \cite{li2017feature} provides a useful categorization of feature selection methods along several dimensions, such as supervised and unsupervised methods; wrapper, filter, and embedded methods; and static or streaming methods. In \cite{di2021supervised}, the authors evaluate several supervised feature selection methods on security datasets. They investigate the prediction performance and computing time of these algorithms. They find that the selected feature set varies with the type of attack recorded in the dataset. 

Most feature selection methods described in the literature have been developed for offline learning where all data fits into memory. In recent years however, \emph{online feature selection} methods have received increasing attention. In this case, the input to the algorithm is either a stream of feature vectors (e.g. \cite{zhou2017online}\cite{alnuaimi2019streaming}\cite{wu2012online}\cite{perkins2003online}\cite{hoi2012online}\cite{sekeh2019geometric}) or a stream of samples (e.g. \cite{wang2013online}\cite{shao2016online}\cite{sekeh2019geometric}). Interestingly, most published online feature selection methods fall into the first category and process a stream of feature vectors. Such methods are not suitable for the problem we study in this paper, since, in our case, the data becomes gradually available as time progresses. Our interest thus is in methods that process streams of samples. Examples of such works are \cite{wang2013online}, \cite{shao2016online}, and \cite{sekeh2019geometric}. 

In \cite{wang2013online}, the authors perform online supervised feature selection as a part of training the perceptron algorithm. During a training step, the values of the selected features are used to update the weights of the perceptron. Features with large weights are selected with high probability and features with small weights are selected with low probability. More closely related to our work is \cite{shao2016online}, which presents an unsupervised online feature selection algorithm based on clustering. The algorithm uses constant memory. In \cite{sekeh2019geometric}, the authors introduce an online feature selection algorithm called Geometric Online Adaption (GOA). A characteristic of this algorithm is that it works for both streams of samples and streams of feature vectors.

Note that a large part of the online feature selection algorithms we found in the literature require the number of features to be selected as an input parameter. The algorithm we present in this paper attempts to find a small but sufficiently large number of features that form a stable feature set.

The authors of \cite{khaire2019stability} provide a recent survey of work on the \emph{stability} of feature selection algorithms. By stability, they mean that an algorithm selects similar feature sets from similar sample sets. Another survey is given in \cite{nogueira2017stability}, which additionally proposes a novel metric for measuring feature set stability. We are using this metric in Section \ref{sec:Computing_stable_feature_sets}.










\section{Conclusions and future work}
\label{sec:conclusions}

In this paper, we  introduced  OSFS,  an  online  algorithm that selects a small set from a large number of available data sources,  which  allows for rapid,  low-overhead learning and prediction. OSFS is instantiated with a set of options and performs a grid search that terminates when a stable feature set has been identified. 

For all traces investigated, we found that OSFS achieves a massive reduction in the size of the feature set while maintaining its performance for training predictors. We have shown the effectiveness of OSFS as an online feature selection algorithm which is robust regarding the sample interval used for feature selection. In addition, we found that, if concept drift occurs, its effect can be mitigated by recomputing the feature set with OSFS and retraining the prediction model.

Regarding future work, there are many ways our method can be improved or extended. We can evaluate OSFS with additional instantiations of feature ranking algorithms, stability conditions, and search policies. Also, the tradeoffs between the three metrics $k$, $t_k$, and the prediction error warrant further study. 

Online learning is only one part of efficiently producing a learning model using online techniques. Other parts include online sample selection \cite{villacca2021online}, and the integration of online feature selection, online sample selection and online training into a joint framework. Finally, we believe it should be investigated how such a framework can be distributed, since we envision both training and prediction as distributed processes in future networked systems.

\section{Acknowledgements}
\label{sec:ack}
The authors are grateful to Andreas Johnsson, Hannes Larsson, and Jalil Taghia with Ericsson Research for fruitful discussion around this work, as well as to Forough Shahab Samani, Kim Hammar, and Rodolfo Villa\c{c}a for comments on an earlier version of this paper. This research has been partially supported by the Swedish Governmental Agency for Innovation Systems, VINNOVA, through project AutoDC, and by Digital Futures through project Democritus.

\bibliographystyle{IEEEtran}
\typeout{}
\bibliography{cnsm_extended}

\end{document}